\documentclass[preprint,12pt]{elsarticle}

\usepackage{amssymb}
\usepackage{booktabs}
\usepackage{amsmath}
\usepackage{graphicx}
\usepackage{diagbox} 
\usepackage{subfigure} 
\usepackage{stfloats}
\usepackage{array} 

\usepackage{url}
\usepackage{float}

\usepackage{lineno}

\journal{review}

\begin{document}

\begin{frontmatter}



\title{HopGAT: Hop-aware Supervision Graph Attention Networks for Sparsely Labeled Graphs}


\author[a]{Chaojie Ji}
\ead{cj.ji@siat.ac.cn}
\author[a]{Ruxin Wang}
\ead{rx.wang@siat.ac.cn}
\author[a]{Rongxiang Zhu}
\ead{rx.zhu@siat.ac.cn}
\author[a]{Yunpeng Cai\corref{cor1}}
\ead{yp.cai@siat.ac.cn}
\author[a]{Hongyan Wu\corref{cor1}}
\ead{hy.wu@siat.ac.cn}

\cortext[cor1]{Corresponding authors at: Shenzhen Institutes of Advanced Technology, Chinese Academy of Sciences, Shenzhen, China.}
\address[a]{
Shenzhen Institutes of Advanced Technology, Chinese Academy of Sciences, Shenzhen, China
}

\begin{abstract}
Due to the cost of labeling nodes, classifying a node in a sparsely labeled graph  while maintaining the prediction accuracy deserves attention. The key point is how the algorithm learns sufficient information from more neighbors with different hop distances.
This study first proposes a hop-aware attention supervision mechanism for the node classification task. A simulated annealing learning strategy is then adopted to balance two learning tasks, node classification and the hop-aware attention coefficients, along the training timeline. Compared with state-of-the-art models, the experimental results proved the superior effectiveness of the proposed Hop-aware Supervision Graph Attention Networks (HopGAT) model. Especially, for the protein-protein interaction network, in a 40\% labeled graph, the performance loss is only 3.9\%, from 98.5\% to 94.6\%, compared to the fully labeled graph. Extensive experiments also demonstrate the effectiveness of supervised attention coefficient and learning strategies.
\end{abstract}




\end{frontmatter}


\section{Introduction}
Node classification is used to predict the class of unlabeled nodes given a partially labeled graph. Node classification is one of the most important applications in analyzing graphs in various areas, including document classification in social science \cite{conf/nips/AtwoodT16}, disease prediction in bioinformatics \cite{journals/corr/abs-1906-05017}, and department classification of an employer in communication networks \cite{ZHANG2019363, ZHANG201938}. However, labeling nodes for a training task is time consuming and sometimes very expensive \cite{LiTowards}. For example, to initially acquire the disease label of a group of disease-causing genes, one must sequence sufficient patient and normal samples \cite{Nguyen2012Detecting}. Being able to predict a node class in a sparsely labeled graph  while maintaining the prediction accuracy deserves more attention \cite{DORNAIKA2019285}.

To address insufficiently labeled data, researchers widely adopt semi-supervised learning, in which both labeled  and  unlabeled neighbor nodes are utilized \cite{Yang2016Revisiting, KIM2019191}. A general assumption in homogeneous graph network research is that a node usually possesses more similar information to its immediate neighbors \cite{ZHANG2019308}. These studies learn node representations on the surrounding nodes or link information via the convolution operation \cite{conf/nips/DefferrardBV16, li2019deepchemstable}. Furthermore, graph attention networks (GATs) \cite{Velickovic:2018we} and various variants \cite{fan2019structured, liu2019hpgat} have been proposed to quantify the closeness of node pairs through attention \cite{journals/corr/abs-1807-07984}.

\begin{figure}[H]
 \centering  \includegraphics[scale=1.0]{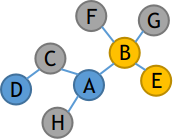}
 \caption{The classification accuracy highly depends on the four colored labeled nodes.}
 \label{graph_struct_data}
\end{figure}

However, the deep learning algorithms \cite{zhang2018graph, wu2018dgcnn}, such as Graph Convolutional Networks (GCNs) \cite{conf/iclr/KipfW17, Bruna:2014vg} and GATs \cite{Velickovic:2018we, doi:10.1021/acs.jmedchem.9b00959},  strongly depend on the labeled nodes to train a prediction model, and thus, the performance is limited by the scale of the labeled data. As seen in Figure \ref{graph_struct_data}, colored nodes are labeled, while gray nodes are unlabeled. To predict the labels of Node C, the algorithms must fully take advantage of the information from Nodes A, B, D and E. Therefore, the prediction accuracy strongly depends on the four labeled color nodes.

The number of hops, namely, the hop value, is adopted to describe the distance or neighborhood relationship between two nodes in the graph network.
The general assumption is that neighbors with different hop values have different influences on their center node.
For classification tasks on sparsely labeled graphs, in which a center node has very few labeled neighbors, it is essential to learn more information from more neighbors with different hop values.
The key point is how the algorithm learns sufficient semantic information from neighbors with different hop values.

\subsection{Motivation}

In this study, we make two key observations.

\textbf{Observation 1: The class labels of neighboring nodes with a smaller hop value are more likely to be consistent with those of their center nodes in the homogeneous graph network.}

\begin{figure}[t]
 \centering  \includegraphics[scale=0.5]{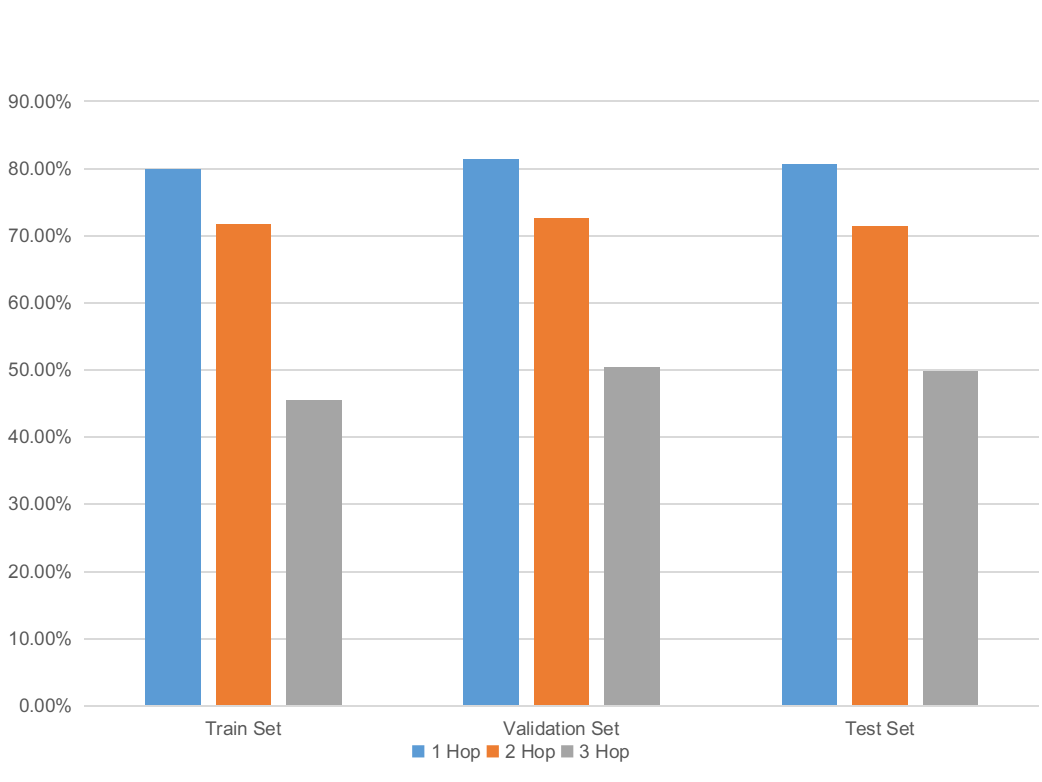}
 \caption{Relation between hop value and class label.}
 \label{class_consistency_distribution}
\end{figure}

We examined the data set Cora \cite{IofSze15}, which is usually used in scientific publication classification tasks, and recorded the label consistence rate, the proportion of neighboring nodes having the same class label as their center nodes. We recorded this consistency rate under different sets of neighboring nodes with a given hop value. In Figure \ref{class_consistency_distribution}, the x-axis is the given hop value, and the y-axis is the label consistency rate.

From the visualization, we can observe that the neighboring nodes with a smaller hop value are more consistent with their center nodes. This is consistent with the general assumption in homogeneous graph networks that a node is more likely similar to its immediate neighbors. This experiment further extends the assumption to neighbors with larger hop values and proves the rationality of introducing hop values to discriminate the closeness of two nodes.

\textbf{Observation 2: General attention models are unable to automatically learn sufficient semantic information from neighbors with different hop values.}

Based on the motivation from Observation 1, we checked the attention coefficients, which should be different between nodes with different hop values since the neighboring nodes with a smaller hop value are more consistent with their center nodes.

\begin{figure}[t]
 \centering
 \includegraphics[scale=0.35]{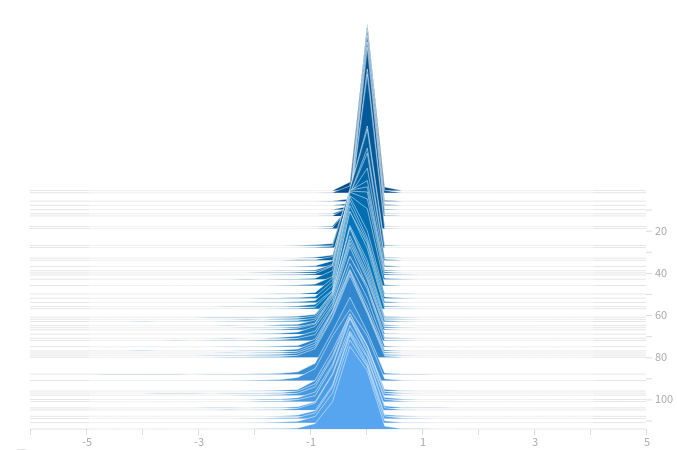}
 \caption{Attention coefficient distribution produced from a trained GAT.}
 \label{attention_coef_distribution}
\end{figure}

We trained a GAT on the Cora citation dataset with the hyper-parameter settings mentioned in \cite{Velickovic:2018we}. The max hop value is set to 2.
All attention coefficients produced by the GATs during the different training epochs are visualized in Figure \ref{attention_coef_distribution}.
In this figure, the horizontal axis shows the value of the attention coefficients, the vertical axis records the occurrence number of the coefficients, and the z-axis represents the training epochs.
This figure illustrates a Gaussian distribution with only one peak. However, we expect a distribution that has multiple peaks or clear boundaries corresponding to different hop values.
This implies that unsupervised attention mechanisms are generally unable to automatically learn the correlation or semantic information for nodes with different hop values in graph networks. This observation is consistent with the work \cite{Knyazev2019Understanding} in that attention should be supervised under different conditions.

\subsection{Contributions}

In this study, we suggest a HopGAT model to address node classification on a sparsely labeled graph. Compared with the general node classification error, we jointly supervise hop-aware attention coefficients in the loss function. Our contributions are described in the following:

\begin{itemize}
  \item To the best of our knowledge, this paper is the first study to propose a hop-aware attention supervision for the node classification task.
  \item We encoded hop values and embedded them into the graph nodes. Subsequently, the graph nodes were simultaneously encoded with semantics and graph structure information.
  \item We proposed a simulated annealing strategy to simultaneously balance two learning tasks, node classification and attention coefficients, along the training timeline.
  \item Experimental results prove the effectiveness of the proposed HopGAT model over the state-of-the-art baselines and quantify our improvement on the sparsely labeled datasets. In addition, the extensive experiments also show the effectiveness of supervised attention coefficients and learning strategy.
\end{itemize}

\section{Related Work}
\subsection{Weakly-supervised Learning in Graph}
Compared with traditional supervised learning, weakly supervised learning aims to address the situation in which precise or sufficient labels are unachievable \cite{journals/corr/CarbonneauCGG16, journals/tnn/FrenayV14}. In the graph domain, semi-supervised learning is widely used to address incomplete labels in a graph, as in \cite{SHI2019518, Xu:2019ty}. These studies mainly focus on graph representation. Vashishth et al. proposed ConfGCN to introduce the concept of the locality of labels \cite{conf/aistats/VashishthYBT19}. They used additional label distribution and co-variance matrices derived from limited labeled nodes. None of these studies took advantage of implicit information in unlabeled neighbors and their hop values.
Yang et al. proposed a method whereby node embedding is trained to predict the class label \cite{Yang2016Revisiting}. They used neighbor context to train the node embedding similar to a Skipgram-like model. The coefficients between two nodes are not computed explicitly or trained in the prediction model.
Inspired by the locality of labels in ConfGCN and neighbor context in Yang's methods, we conducted a further analysis of the changing of the node labels with different hop values. Based on the observation that neighboring nodes with smaller hop values are more likely to be consistent with their center nodes, we embed the hop value information in the classification function in this study.

\subsection{Attention in Graph}
Given the trend whereby convolution operations are being generalized into arbitrary graph networks, more effective methods to aggregate neighboring nodes and locate ``closest'' nodes from center nodes are desired, as in \cite{lee2018graph, wang2018brief}.
An attention mechanism aids a model to ``focus on the most relevant parts of the input to make decisions'' \cite{journals/corr/abs-1803-02155, vaswani2017attention}.
Although attention is widely used in locating the closest nodes and have achieved state-of-the-art performance \cite{shanthamallu2019gramme}, the studies in \cite{conf/kdd/ChoiBSSS17} showed that attention should be supervised to obtain a better performance under different conditions. In this study, we jointly supervise the hop-aware attention coefficients and the node classification error in the loss function to better train an algorithm given insufficiently labeled nodes.

\subsection{Supervision on Attention in Graph}
To better understand attention mechanism in graph convolution network, Knyazev et al. proposed ChebyGIN and imposed a supervision on attention coefficients as a controlled environment \cite{Knyazev2019Understanding}.
They noticed the influence of attention and mentioned that the accuracy of a model could depend exponentially on attention correctness.
The proposed mechanism of attention supervision is goal-directed which can't be directly applied in other tasks, such as node classification.

\section{Preliminary}
\label{section:Preliminary: Graph Attention Networks}
Before introducing our proposed method, we provide a brief overview of the semi-supervised GATs which are composed of several single graph attentional layer. We only describe a single graph attention layer here.

Given a graph, the input to current layer are defined as node features, i.e. $h=\{h_1, h_2,\cdots, h_n\}$, where $n$ is the number of nodes.

The attention coefficients between two nodes can be calculated as follows:
\begin{equation}
  \label{Preliminary: Graph Attention Networks:E1}
  \alpha_{ij} = \frac{exp(a(h_i, h_j))}{\sum_{k\in \mathcal{N}_i}exp(a(h_i, h_k))}
\end{equation}
where $exp$ is the exponential function, $\mathcal{N}_i$ is the neighboring nodes of $i$ in the graph, and $a$ is a function used to estimate the importance of one node to another.

Once obtained, the attention coefficients of node $i$ are used to compute a linear combination of the features corresponding to its neighboring nodes, which can be considered as the updated output features $h'_i$ for node $i$:
\begin{equation}
  \label{Preliminary: Graph Attention Networks:E2}
  h'_i = \sigma (\sum_{j \in \mathcal{N}_i}\alpha_{ij}Wh_j )
\end{equation}
where $W$ is a weight matrix.

Similar to the Transformer \cite{vaswani2017attention}, multi-head attention is employed in GATs. $K$ independent attention mechanisms execute a transformation as in Equation \ref{Preliminary: Graph Attention Networks:E2}. Then, the produced features are concatenated as
\begin{equation}
  \label{Preliminary: Graph Attention Networks:E3}
  h'_i = {||}^K_{k=1} \sigma (\sum_{j \in \mathcal{N}_i}\alpha^k_{ij}W^kh_j )
\end{equation}
where $||$ denotes the concatenation operation, and $k$ corresponds to the $k$th head.

For the final layer, GATs employ averaging and nonlinearity to a specific task:
\begin{equation}
  \label{Preliminary: Graph Attention Networks:E4}
  h'_i = \sigma (\frac1K \sum_{k=1}^K \sum_{j \in \mathcal{N}_i}\alpha^k_{ij}W^kh_j )
\end{equation}

\section{Method}
\begin{figure*}[t]
  \centering
  \includegraphics[scale=0.47]{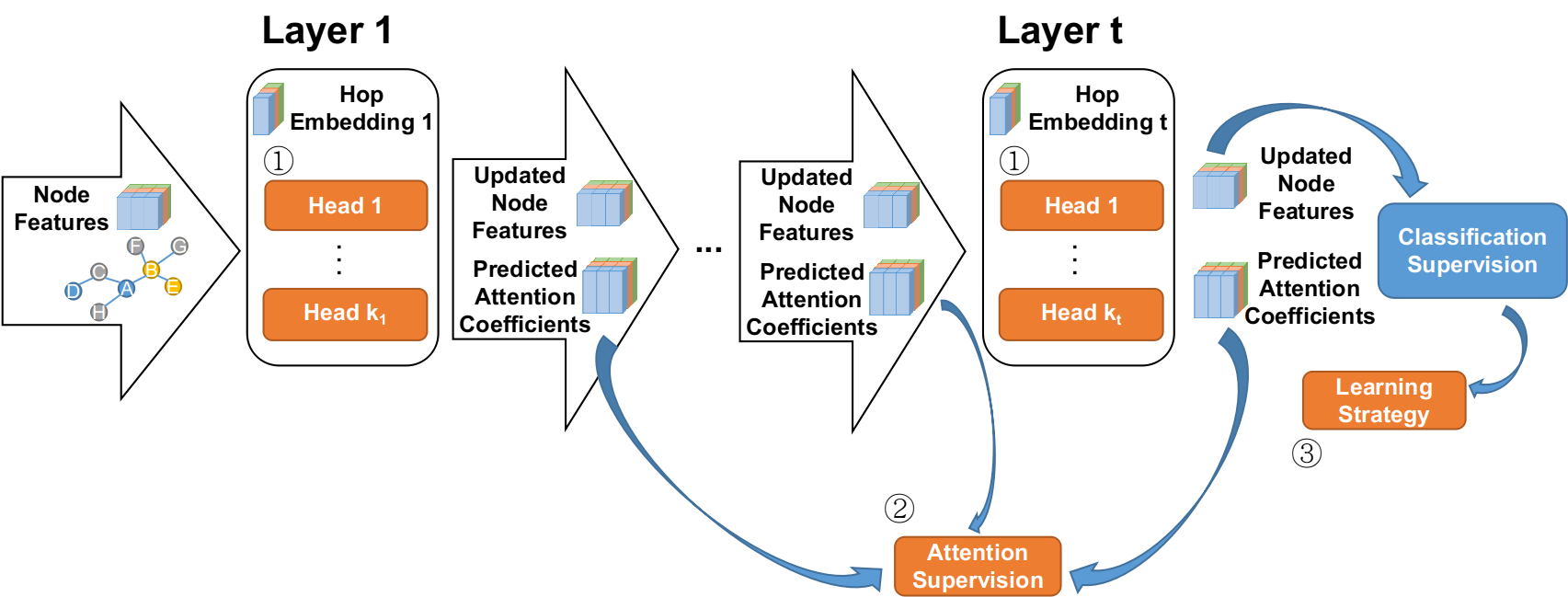}
  \caption{Overall framework of an Hop-GAT.}
  \label{overall_framework}
\end{figure*}

In this section, we describe the architecture of our model - HopGAT. The model has three main components: a hop encoding and attention mechanism, attention supervision, and a learning strategy, as shown in Figure \ref{overall_framework}. The other modules shown in the figure are inherited from the GATs.

In component 1, we first encode the hop value into a vector and embed the vector into each node feature. Then, we calculate the attention coefficients of each center node to its neighboring nodes with hop information. Finally, we apply multi-head hop attention for each layer, which can draw different features of the graph network. In the next layer, the node features are updated according to features from the previous layer, and the attention coefficients for each node pair are delivered to the attention supervision. In component 2, all the attention coefficients from every head are collected. The gap between the computed coefficients and the defined ground-truth coefficients are summed to form an attention loss, which will be one part of the total loss function used to train the prediction model. In the last layer, the node classification loss is calculated. Once the attention loss and node classification loss are obtained, our learning strategy is performed in component 3 to balance these two types of losses during the training procedure.

\subsection{Attention Mechanism}
\subsubsection{Hop Encoding}
Existing attention-based graph methods usually specify the maximum hop value between a center node and its neighbors, e.g. 1, in GATs. Then, the numerical hop value in a graph is coarsely equivalent to a Boolean variable representing whether a node is a neighbor of another node. However, as we have noted in Observation 1, the class labels of neighboring nodes with a smaller hop value are more likely to be consistent with those of their center nodes in a homogeneous graph network, and the hop value offers more information than a Boolean variable.

In this study, a hop value is used to express the closeness or similarity of a node to its center node. Therefore, we encode a hop value into a $d$-dimensional vector as $he$:
\begin{equation}
  \label{Method:Attention Mechanism:E0}
  inv_i = exp(i * (-\frac{log(max_{hv})}{max(\frac d2-1, 1)}))
\end{equation}

\begin{equation}
\label{Method:Attention Supervision:E1}
he_{hv, i}=\left\{
\begin{aligned}
sin(hv * inv_i), & & i <= d/2 \\
cos(hv * inv_i), & & i > d/2 \\
\end{aligned}
\right.
\end{equation}

where $max_{hv}$ is the pre-defined maximal hop value, $max(\cdot)$ is a maximum value function, $d$ is the dimension of the hop embedding, $i$ is the index of the dimensions, $hv$ is the corresponding hop value, $sin$ and $cos$ are sine and cosine functions, respectively, and $he_{hv}$ represents the hop encoding for the hop value $hv$.

Through this definition, not only the absolute hop value but also the relative hop value can be learned since, as a fixed offset $p$, $he_{hv + p}$ can be transformed as a linear function of $he_{hv}$. Since the number of heads in each layer can be different, the dimensions of nodes in each hidden layer are also different. Therefore, the dimension of the hop embedding in each layer may be different.

\subsubsection{Attention Mechanism}
We propose two attention mechanisms based on the above hop encoding: product-based attention and addition-based attention.

For a node pair $(i, j)$, $i$ and $j$ are the center and neighboring nodes, respectively. $hv_{i,j}$ is the hop value from $i$ to $j$. The hop encoding for the $t$th layer is written as $he^{t}$, where each line represents a hop encoding for a hop value.

\begin{itemize}
  \item \textbf{Product-based Attention}
  This attention mechanism is mainly based on the dot product between the embeddings of two nodes.

  \begin{equation}
  \label{Method:Attention Mechanism:E2}
  e_{ij} = a_c(h_i) \odot (a_n(h_j) + lookup(a_{he}(he^{t}), hv_{i, j}))
  \end{equation}
  $e_{ij}$ indicates the importance of node $j$ to node $i$, $a_c$, $a_n$ and $a_{he}$ are three independent single-layer feedforward neural networks, $\odot$ denotes the dot product operation, and $lookup$ is a function that looks up the hop encoding for the hop value $hv_{i,j}$ in $he^{t}$.

  \item \textbf{Addition-based Attention}
  This mechanism is based on the addition operation on two node embeddings.
  The attention coefficients are formulated as follows:
\begin{equation}
    \label{Method:Attention Mechanism:E4}
    \begin{split}
    e_{ij} = LeakyReLU(lookup(a_{he}(he^{t}), hv_{i, j})    \\
    \odot (a_c(h_i) || a_n(h_j)))
    \end{split}
  \end{equation}

  where the symbol $||$ is a concatenation operation. The LeakyReLU activation function is used to execute the non-linearization operation.
\end{itemize}

To make coefficients easily comparable, we normalize them across all neighbors $j$ using the softmax function:

\begin{equation}
\label{Method:Attention Mechanism:E5}
\alpha_{ij} = softmax_j(e_{ij}) = \frac{exp(e_{ij})}{\sum_{k\in \mathcal{N}_i}exp(e_{ik})}
\end{equation}
where $\mathcal{N}_i$ is the neighboring node set of node $i$. Once $\alpha_{ij}$ is obtained, we use Equations \ref{Preliminary: Graph Attention Networks:E3} - \ref{Preliminary: Graph Attention Networks:E4} to perform the final node classification.

The hop value of self-connection is defined as 0, and it is also encoded into a non-zero hop encoding vector.
Through this design, all attention coefficients are uniformly calculated according to the node features and hops.

\subsection{Attention Supervision}
\subsubsection{Ground-truth Attention}
The hop value represented in the graph data is a label-free and effective indicator for quantifying the correlation between two nodes. Specifically, the correlation between two directly connected nodes in the graphs should be assigned a larger value, while the coefficient between two indirectly connected or disconnected nodes is expected to have a smaller value. In addition, a prior investigation showed that ``classification accuracy depends exponentially on attention correctness'' \cite{Knyazev2019Understanding} and that the general attention models cannot effectively automatically learn sufficient semantic information from neighbors with different hop values. All of this motivates us to develop a hop-aware approach to supervise the training process for attention coefficients between two nodes.

Generally, the correlation between two nodes becomes weaker when their hop value is over a certain value. Therefore, we set a boundary for the hop value. The coefficients between the center node and its neighboring nodes are uniformly grouped into a default value when the hop value reaches the maximum value, $max_{hv}$, which means a weak connection between two nodes.
We formulate the definition of ground-truth attention $e^{GT}$ as follows:

\begin{equation}
\label{Method:Attention Supervision:E1}
e^{GT}_{ij}=\left\{
\begin{aligned}
1, & & hv_{i, j} = 0 \\
1- hv_{i, j}, & & 0 < hv_{i, j} < max_{hv} \\
1- max_{hv},  & & hv_{i, j} >= max_{hv}
\end{aligned}
\right.
\end{equation}
This can be interpreted as follows: the larger the hop value is, the smaller the ground-truth attention.

When the hop value is greater than two, a negative ground-truth attention value will be assigned. This negative value does not indicate a negative association between these two nodes. This coefficient has not been normalized by the softmax function as in Equation \ref{Method:Attention Mechanism:E5}. After normalization, the attention coefficients will be between 0 and 1, indicating the strong and weak correlation between two nodes.

To be more accurate, we use $e^{kl}_{ij}$ to denote the attention coefficient between nodes $i$ and $j$ of the $k$th head in the $l$th layer, which are produced during the training procedure as in Equations \ref{Method:Attention Mechanism:E2} and \ref{Method:Attention Mechanism:E4}.
We use the mean square error to control the distance between the ground-truth attention $e^{GT}$ and the computed coefficients $e^{kl}_{ij}$.
\begin{equation}
\label{Method:Attention Supervision:E2}
\mathcal{L}_{att} = \frac{\sum_{i=1}^L \sum_{j=1}^K\sum_{l=1}^N \sum_{k=1}^N {(e_{ij}^{GT} - e_{ij}^{kl})^2}}{L * K * N * N}
\end{equation}

$\mathcal{L}_{att}$ will be used as a part of the loss function to supervise the training process for the attention coefficients between two nodes.

\subsubsection{Sample Strategy}
In total, there are $L * K * N * N$ attention coefficients in the calculation of $\mathcal{L}_{att}$.
To decrease the computation cost, especially in a graph with a large number of nodes,
we thus proposed a random sampling strategy, $sample(r)$, with which a subset of node pairs is sampled with sample ratio $r$.


The number of node pairs with $hv_{i,j} >= max_{hv}$ is greater than that with less than $max_{hv}$.
For example, in the Citeseer dataset, there are approximately 12,000 node pairs with less than 2 hops, whereas there is approximately 11,000,000 pairs with more hops. For balancing the distribution between node pairs with hop values of greater or less than $max_{hv}$, $sample(r)$ only samples from the node pairs with hop values greater than $max_{hv}$. Furthermore, we sample each batch differently to guarantee the diversity of the training data. $\mathcal{L}_{att}$ is calculated as follows:

\begin{equation}
\label{Method:Attention Supervision:E3}
\mathcal{L}_{att} = \frac{\sum_{(i,j) \in sample(r)} \sum_{l=1}^L \sum_{k=1}^K {(e_{ij}^{GT} - e_{ij}^{kl})^2}}{L * K * count(sample(r))}
\end{equation}

\subsection{Learning Strategy}
In addition to the loss function $\mathcal{L}_{att}$ used to supervise the attention coefficients, we also include the general node classification loss to measure the classification error.

The final objective for optimization is the linear combination of these two terms.
\begin{equation}
\label{Method:Learning Strategy:E1}
\mathcal L = (1 - \gamma) \mathcal L_{cls} + \gamma \mathcal L_{att}
\end{equation}
where $\gamma$ is used to find a balance between node classification and attention supervision losses.

Inspired by the analysis that attention coefficients are more likely to be imprecise at the beginning of training \cite{Knyazev2019Understanding}, more powerful supervision of the attention should be imposed at the early stage of the training. From another perspective, $\mathcal L_{cls}$ is a ``strong'' label and is closely related to the final task goal, i.e., node classification, while $\mathcal L_{att}$ is auxiliary. Thus, a consideration of balancing them with time is necessary. A simulated annealing procedure is adopted to help the model find the best combination of these two parts of the total loss function.

We first define the transformation of the temperature along the training time:

\begin{equation}
\label{Method:Learning Strategy:E2}
temp^t=\left\{
\begin{aligned}
temp_{ini}, & & t = 0\\
temp^{t-1} * \epsilon, & & temp^{t-1} * \epsilon >= temp_{fin}\ and\ t > 0\\
temp^{t-1}, & & temp^{t-1} * \epsilon < temp_{fin}\ and\ t > 0\\
\end{aligned}
\right.
\end{equation}

where $temp_{ini}$ and $temp_{fin}$ are the initial and final temperatures, respectively; $temp^t$ indicates the temperature at the $t$th time step; $\epsilon$ is the decay rate; and $temp_{ini}$, $temp_{fin}$ and $\epsilon$ are all pre-defined hyperparameters.

Then, $\mathcal L_{cls}$ and $\mathcal L_{att}$ are biased through $\gamma$ along the training time as follows:
\begin{equation}
\label{Method:Learning Strategy:E3}
\gamma=\left\{
\begin{aligned}
min(exp(-\frac{\frac{1}{\mathcal L_{att}}}{temp^t}), \gamma_{str}), & temp^{t-1} * \epsilon < temp_{fin}\\
exp(-\frac{\frac{1}{\mathcal L_{att}}}{temp^t}), & otherwise
\end{aligned}
\right.
\end{equation}

where $min$ is the minimum function. $\gamma_{str}$ is a hyperparameter designed to prevent a sharp increase of $\mathcal L_{att}$ at the tail of the training process. The learning procedure will be explained in greater detail in the experiment section.

$\mathcal{L}_{cls}$ measures the classification error and is applied to all the labeled nodes during the training procedure. If the labeled nodes in the graph network are insufficient, the trained model will not be sufficiently accurate. In this study, the hop value is introduced as an effective supplementary to the insufficient labels to train the model but without increasing the labeling cost.

\section{Experiments}
In this section, we evaluate the performance of the proposed hop-aware attention supervision model to address the node classification task on a sparsely labeled graph network. We also investigate the effectiveness of the supervised attention coefficients and our learning strategy.

\subsection{Dataset}

This experiment includes two types of tasks: inductive learning and transductive learning.
For the semi-supervised tasks, if the unlabeled test nodes do not participate in the training procedure, we call this task inductive learning, whereas if the unlabeled test data are all observed and utilized during the training phase, the task is transductive learning.

Cora, Citeseer and PubMed are chosen as our benchmark datasets for transductive tasks \cite{IofSze15}. In all of these datasets, nodes denote documents, and edges correspond to citation relations. Node features are elements of a bag-of-words representation of a document. The task is to predict the unique document class among multiple documents. The protein-protein interaction (PPI) dataset is used to evaluate inductive tasks as in \cite{sen:aimag08}. Each node is a protein. Positional and motif gene sets and immunological signatures are used to represent a protein. It is a multi-label task that simultaneously predicts multiple protein functions (labels). We used the preprocessed data from \cite{conf/nips/HamiltonYL17} in our experiments.

We evaluate how the proposed algorithm works on the insufficiently labeled graph network. Therefore, we change the proportions of the labeled nodes in the training set. We then reorganize all the datasets.

\begin{table*}[!htbp]
 \centering
 \scriptsize
 \begin{tabular}{c|ccccccccccc}
   \hline
   \multicolumn{1}{c|}{}&\#Nodes&\#G&\#F&\# C&\#VN&\#TN&\multicolumn{5}{c}{\# Training Nodes with Label Rate (\%)}\\
   \multicolumn{1}{c|}{}& & & & & & &20&40&60&80&100\\
   \hline
   Cora & 2708 & 1 & 1433 & 7 & 500 & 1000 & 242 & 484 & 725 & 967 & 1208\\
   Citeseer & 3327 & 1 & 3703 & 6 & 500 & 1000 & 363 & 725 & 1008 & 1450 & 1812\\
   PubMed & 19717 & 1 & 500 & 3 & 500 & 1000 & 3644 & 7287 & 10931 & 14574 & 18217\\
   PPI & 56944 & 24 & 50 & 121 & 6514 & 5524 & 8982 & 17963 & 26944 & 35925 & 44906\\
   \hline
 \end{tabular}
 \caption{Number of randomly sampled nodes. Column \#G, \#F, \#C, \#VN and \#TN denote the number of graphs, features, classes, validation nodes and test nodes, respectively.}
 \label{dataset_overview}
\end{table*}

The validation and test sets completely follow the experimental setup of \cite{Velickovic:2018we}. In all transductive tasks, 500/1000 nodes serve for the validation/test sets; 6514/5524 nodes (in 2/2 graphs) are used for the validation/test sets in the inductive tasks. The remaining nodes are placed into the training set. Then, we randomly sample nodes at rates of 20\%, 40\%, 60\%, 80\%, and 100\% without replacement from the training set for each dataset. We reserve the labels of the sampled nodes but mask the labels of other nodes. Subsequently, we obtained 5 variants for each dataset. The details of the number of nodes, classes, graphs and features in each dataset are listed in Table \ref{dataset_overview}.

\subsection{Experimental Setup}

We compared the proposed HopGAT against state-of-the-art methods. GATs \footnote{\url{https://github.com/PetarV-/GAT}}, GCNs \footnote{\url{https://github.com/tkipf/gcn}} \cite{conf/iclr/KipfW17} and ConfGCN \footnote{\url{https://github.com/malllabiisc/ConfGCN}} \cite{conf/aistats/VashishthYBT19} were chosen for the transductive tasks. GraphSAGEs\footnote{\url{https://github.com/williamleif/GraphSAGE}} \cite{conf/nips/HamiltonYL17} and GATs were chosen for the inductive tasks. Furthermore, we selected different variants of GraphSAGEs, i.e., GraphSAGE with the mean-based aggregator (Mean), LSTM-based aggregator (Seq), max-pooling aggregator (Maxpool), mean-pooling aggregator (Meanpool) and GCN-based aggregator (GCNagg).
For each task, we run the same experiment five times and record the average performances and the standard deviations. For the inductive tasks, we used the metric of the micro-averaged $F_1$ score and the accuracy for the transductive tasks instead.

\begin{table*}[!htbp]
 \scriptsize
 \centering
 \begin{tabular}{c|cccccccccc}
   \hline
   \multicolumn{1}{c|}{}&$dp_1$&$dp_2$&$dp_3$&$L_2$&Attention&\# L&\# Heads&\# Features&BS&LR\\
   \hline
   Cora & 0.2 & 0.0 & 0.2 & 0.0001 & Addition & 2 & [8,1] & [8,7] & 1 & 0.005\\
   Citeseer & 0.6 & 0.2 & 0.6 & 0.0 & Addition & 2 & [8,1] & [8,6] & 1 & 0.005\\
   PubMed & 0.0 & 0.0 & 0.0 & 0.0 & Addition & 2 & [8,8] & [8,3] & 1 & 0.01\\
   PPI & 0.0 & 0.0 & 0.0 & 0.0 & Product & 3 & [4,4,6] & [256,256,121] & 2 & 0.005\\

   \hline
 \end{tabular}
 \caption{Common Hyperparameters of GATs and HopGAT on different tasks. Column \#L, \#BS and \#LR denote the number of layers, batch size and learning rate, respectively.}
 \label{hyperparameters}
\end{table*}

\begin{table}[!htbp]
  \scriptsize
  \centering
  \fontsize{9}{9}\selectfont
  \caption{Hyperparameters of GCNs and ConfGCNs in transductive tasks.}
  \label{tab:hyperparmeters_GCN_ConfGCN}
    \begin{tabular}{ccccccc}
    \toprule
    { }{ }{ }&
    \multicolumn{3}{c}{GCN}&\multicolumn{3}{c}{ConfGCN}\cr
    \cmidrule(lr){2-4} \cmidrule(lr){5-7}
    &Cora&Citeseer&PubMed&Cora&Citeseer&Pubmed\cr
    \midrule
    dp&0.1&0.4&0.0&0.8&0.3&0.0\cr
    $L_2$&1e-4&0.0&0.0&0.01&0.05&0.0\cr
    \bottomrule
    \end{tabular}
    \label{tab:transduvtive_parameters}
\end{table}

We applied dropout \cite{srivastava2014dropout}, skip connections \cite{he2015residual} and $L_2$ regularization techniques to alleviate over-fitting.
For each layer in the HopGAT model, we applied three dropout units: when receiving the updated node representation from the previous layer, archiving the normalized attention coefficients  in Equation \ref{Preliminary: Graph Attention Networks:E2}, and obtaining the transformed node representation in Equation \ref{Preliminary: Graph Attention Networks:E2}. We denote these dropout rates as $dp_1$, $dp_2$ and $dp_3$, respectively. The skip connections were employed across the intermediate layers when the number of layers was greater than two.
An exponential linear unit (ELU) \cite{ClevertUH15} instead of a $\sigma$ function in Equation \ref{Preliminary: Graph Attention Networks:E2} is implemented when the computation does not occur in the last layer. We applied a single-layer feedforward neural network when the dimension of the inputs is not equal to the number of features in the last layer. We fixed the number of output units in the single-layer feedforward neural networks to 1, i.e., $a_c$, $a_n$ and $a_{he}$ in Equation \ref{Method:Attention Mechanism:E2}.
The common parameters of GATs and HopGATs are presented in detail in Table \ref{hyperparameters}. We mainly adjusted the dropout rate and $L_2$ regularization for GCN and ConfGCN, which are listed in Table \ref{tab:transduvtive_parameters}. The coefficient of $L_{smooth}$ in ConfGCN's objective function was set to 1.0 in the PubMed dataset. The other hyperparameters not mentioned here were all derived from their original publications, i.e., GCN \cite{conf/iclr/KipfW17}, ConfGCN \cite{conf/aistats/VashishthYBT19} and GraphSAGEs \cite{conf/nips/HamiltonYL17}.

We used the Adam SGD optimizer \cite{kingma2014method}. An early stopping strategy is applied to the loss for the node classification and also the accuracy (or micro-F1) on the validation nodes with a patience of 100 epochs in all experiments. For the learning strategy, the initial temperature is set to 100, and the lowest temperature is 1. The decay rate for the Cora dataset is 0.95, and it is 0.85 for the other datasets. The saturation gamma is 0.25.
For the sampling strategy, we used different percentages to sample node pairs - 0.0003, 0.0005 and 0.0001 for Cora, Citeseer and PubMed, respectively.

\subsection{Results}


\subsubsection{Inductive Task}

\begin{table*}[!htbp]
  \scriptsize
  \centering
  \begin{tabular}{c|cccccccc}
    \hline
    \multicolumn{1}{c|}{}&GCNagg&Mean&Meanpool&Maxpool&Seq&GAT&Hop-GAT&Imp\\
    \hline
    PPI(100\%) & 51.6$\pm$0.5 & 58.0$\pm$0.9 & 59.0$\pm$0.5 & 60.2$\pm$0.7 & 61.2$\pm$0.5 & 97.3$\pm$0.2 & \textbf{98.5$\pm$0.1} & +1.2\\
    PPI(80\%) & 45.3$\pm$0.7 & 55.6$\pm$0.2 & 52.8$\pm$1.0 & 52.6$\pm$0.7 & 56.2$\pm$0.6 & 96.8$\pm$0.2 & \textbf{97.8$\pm$0.3} & +1.0\\
    PPI(60\%) & 46.9$\pm$0.4 & 54.4$\pm$0.7 & 52.5$\pm$1.4 & 51.0$\pm$2.3 & 54.0$\pm$0.5 & 95.2$\pm$0.4 & \textbf{96.7$\pm$0.6} & +1.5\\
    PPI(40\%) & 48.8$\pm$0.8 & 52.9$\pm$0.2 & 49.8$\pm$0.4 & 48.8$\pm$0.2 & 52.2$\pm$0.9 & 91.5$\pm$0.4 & \textbf{94.6$\pm$1.8} & +3.1\\
    PPI(20\%) & 44.1$\pm$0.3 & 50.5$\pm$0.4 & 43.2$\pm$1.1 & 43.6$\pm$1.5 & 48.4$\pm$0.7 & 83.3$\pm$0.3 & \textbf{88.6$\pm$1.2} & +5.3\\
    \hline
  \end{tabular}
  \caption{F1 score of inductive tasks. Symbols Imp denotes the improvement.}
  \label{F1_Results}
\end{table*}

The results for the inductive tasks are listed in Table \ref{F1_Results}. The improvement column records the improved values of HopGAT against GATs, which shows better performance compared to other baseline methods.

Compared with GATs, the results demonstrate that HopGATs obtain a significant gain across all datasets with different proportions of labeled nodes.
Specifically, the smallest improvement is 1.0\% for the 80\% label rate, and the highest improvement is 5.3\% for the 20\% label rate. It is observed that given fewer labeled nodes, a higher performance gain is achieved, from 1.2\% to 5.3\% - except for a small fluctuation at the 80\% label rate.
The results show that the proposed model is effective in addressing the inductive task on the sparsely labeled graph network.

Another important observation from the table is that we cannot achieve a proportional performance gain by labeling additional nodes. Labeling more nodes, e.g., from 20\% to 40\%, the performance gains is nearly 6\%.
With a 40\% labeled graph, the performance loss is only 3.9\%, from 98.5\% to 94.6\%, compared to the fully labeled graph.
More labeled nodes results in smaller performance gains. Therefore, if necessary, we should balance the cost of labeling more nodes and performance gain.

\subsubsection{Transductive Tasks}

\begin{table*}[!htbp]
   \scriptsize
   \centering
   \begin{tabular}{c|ccccccc}
     \hline
     \multicolumn{1}{c|}{}& GCN & Conf-GCN &GAT &Hop-GAT &Improvement \\
     \hline
     Cora (100\%)& 87.2 $\pm$ 0.4 & 87.5 $\pm$ 0.4 & \textbf{88.1 $\pm$ 0.3} & \textbf{88.1 $\pm$ 0.4} & + 0.0\\
     Cora (80\%) & 86.8 $\pm$ 0.2 & 87.3 $\pm$ 0.2 & 86.8 $\pm$ 0.2 & \textbf{87.3 $\pm$ 0.4} & + 0.5\\
     Cora (60\%) & 85.8 $\pm$ 0.2 & 86.5 $\pm$ 0.4 & 86.5 $\pm$ 0.2 & \textbf{87.1 $\pm$ 0.1} & + 0.6\\
     Cora (40\%) & 84.5 $\pm$ 0.3 & 86.1 $\pm$ 0.1 & 86.0 $\pm$ 0.4 & \textbf{86.5 $\pm$ 0.3} & + 0.5\\
     Cora (20\%) & 82.4 $\pm$ 0.3 & 83.0 $\pm$ 0.2 & 83.1 $\pm$ 0.3 & \textbf{83.6 $\pm$ 0.2} & + 0.5\\
     \hline
     Citeseer (100\%)& 78.8 $\pm$ 0.2 & 77.5 $\pm$ 0.2 & 78.4 $\pm$ 0.9 & \textbf{79.5 $\pm$ 0.3} & + 1.1\\
     Citeseer (80\%) & 78.0 $\pm$ 0.3 & 76.8 $\pm$ 0.4 & 77.4 $\pm$ 0.9 & \textbf{78.1 $\pm$ 0.4} & + 0.7\\
     Citeseer (60\%) & 76.9 $\pm$ 0.5 & 76.1 $\pm$ 0.7 & 77.0 $\pm$ 0.3 & \textbf{77.8 $\pm$ 0.5} & + 0.8\\
     Citeseer (40\%) & 75.3 $\pm$ 0.3 & 74.9 $\pm$ 0.5 & 75.5 $\pm$ 0.7 & \textbf{76.2 $\pm$ 0.4} & + 0.7\\
     Citeseer (20\%) & 72.8 $\pm$ 0.2 & 74.3 $\pm$ 0.4 & 73.9 $\pm$ 0.3 & \textbf{74.3 $\pm$ 0.1} & + 0.4\\
     \hline
     PubMed (100\%) & 87.3 $\pm$ 0.1 & 85.9 $\pm$ 0.3 & 87.5 $\pm$ 0.7 & \textbf{88.9 $\pm$ 0.2} & + 2.4\\
    PubMed (80\%) & 87.5 $\pm$ 0.2 & 86.1 $\pm$ 0.4 & 87.5 $\pm$ 0.3 & \textbf{88.7 $\pm$ 0.6} & + 1.2\\
     PubMed (60\%) & 86.9 $\pm$ 0.2 & 86.0 $\pm$ 0.4 & 86.9 $\pm$ 0.2 & \textbf{88.3 $\pm$ 0.2} & + 1.4\\
     PubMed (40\%) & 86.5 $\pm$ 0.1 & 86.0 $\pm$ 0.4 & 86.0 $\pm$ 0.4 & \textbf{87.6 $\pm$ 0.2} & + 1.6\\
     PubMed (20\%) & 86.5 $\pm$ 0.2 & 85.1 $\pm$ 0.3 & 86.2 $\pm$ 0.2 & \textbf{87.3 $\pm$ 0.4} & + 1.1\\
     \hline
   \end{tabular}
   \caption{Accuracy of transductive tasks.}
   \label{Accuracy_Results}
 \end{table*}

The results of all transductive tasks are shown in Table \ref{Accuracy_Results}.
Compared with GATs, the maximum improvement for Cora, Citeseer and PubMed are 0.6\%, 1.1\% and 2.4\% respectively. The minimum improvements are 0.0\%, 0.4\% and 1.1\% respectively. This proves the effectiveness of our model on transductive tasks.

We also investigate the improved average performance on individual datasets. The improved average accuracy is 0.42\%, 0.74\% and 1.54\% on the Cora, Citeseer and PubMed datasets, respectively. These datasets include 2708, 3327, and 19717 nodes, respectively. This shows that the larger the graphs are, the more benefit that the proposed model can obtain.

We can observe that the greater the number of labeled nodes, the higher the performance gain obtained on Citeseer and PubMed. This is different from the inductive task, in which fewer labeled nodes results in higher performance gains.
This could be caused by the large difference between the mechanisms of inductive and transductive tasks. In the inductive tasks, validation and test nodes are completely unseen in the entire training process, whereas they directly participate in the training process in the transductive tasks. This means that with the HopGAT model, the training nodes could learn more correlations from the validation/test nodes in the transductive tasks, even when the test nodes are not labeled. In addition, the smaller the dataset is, the stronger the randomness during the sampling, which can cause a fluctuation of the performance gains in certain cases.

\subsubsection{Effectiveness of Supervised Attention Coefficients}

\begin{figure*}[t]
  \centering
  \subfigure[Attention distribution in the 1st layer.]{\includegraphics[width=2.6in]{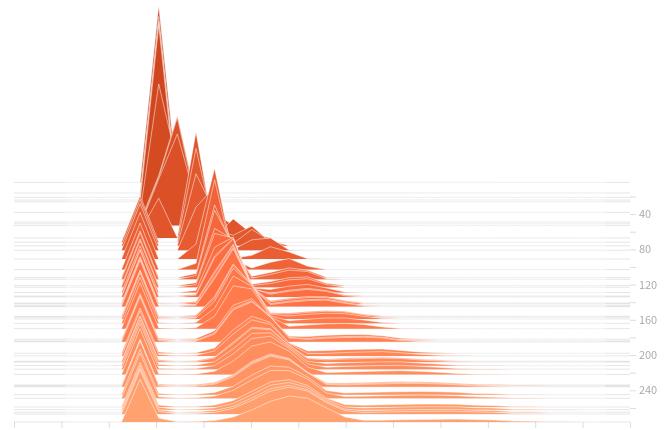}}
  \subfigure[Attention distribution in the 2nd layer.]{\includegraphics[width=2.6in]{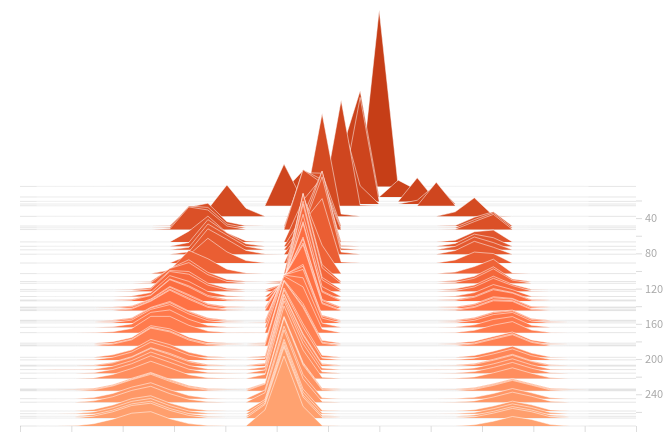}}
  \caption{Attention distribution produced from the trained HopGAT.}
  \label{attention_distribution}
\end{figure*}

We investigate the effectiveness of the supervised attention coefficients. We thus generate a visualization of the attention coefficients produced by the proposed HopGAT model. We trained on the training set of PubMed. We record the distribution of the produced attention coefficients during the different training epochs. Figure \ref{attention_distribution} (a) comes from a head of the 1st layer, and Figure \ref{attention_distribution} (b) is from the 2nd layer. Similar to Figure \ref{attention_coef_distribution}, the horizontal axis shows the value of the attention coefficients, and the vertical axis records the occurrence number of the coefficients. Compared to Figure \ref{attention_coef_distribution}, there are three clusters that correspond to the attention coefficients for 0, 1 and greater than 1 hop value from left to right.

Comparing Figure \ref{attention_distribution} (a) to Figure \ref{attention_distribution} (b), we noticed that the attention coefficients inside the 2nd layer have more clear boundaries among different hop values than those in the 1st layer. This is consistent with the expectation that the signal received from the deeper layer is much stronger than the first layers due to back propagation.

\subsubsection{Effectiveness of Learning Strategy}
In this section, we evaluate the proposed learning strategy. We train the HopGAT model on PPI with a 100\% label rate and the hyperparameters mentioned above. Then, we record the entire training process as a learning curve of the loss for the node classification, attention coefficients, and micro-averaged $F_1$ score.  We also visualize the changes of $\gamma$ with increasing training epochs.

\begin{figure}[h!]
  \centering
  \subfigure[Variation of $\gamma$.]{\includegraphics[width=2.6in]{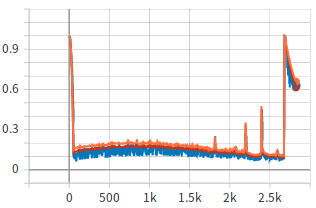}}
  \subfigure[Learning curve between epoch and node classification loss.]{\includegraphics[width=2.6in]{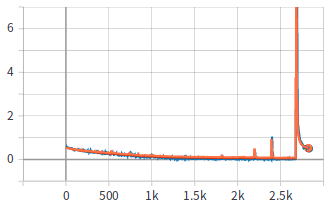}}
  \subfigure[Denoised learning curve between epoch and attention coefficient loss.]{\includegraphics[width=2.6in]{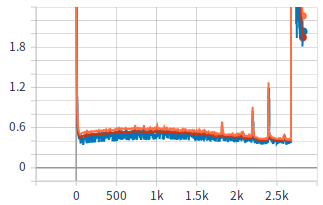}}
  \subfigure[Learning curve between epoch and micro-F1 score.]{\includegraphics[width=2.6in]{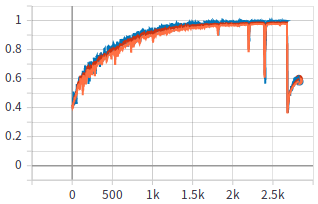}}
  \caption{The curves related to our proposed learning strategy.}
  \label{learning_strategy_curves}
\end{figure}

The value of $\gamma$ is used to balance the node classification and attention supervision loss functions. As shown in Figure \ref{learning_strategy_curves} (a), at the early stage of training, i.e., the first 100 steps, more powerful supervision on the attention was imposed due to a larger $\gamma$. Once $\mathcal L_{att}$ is relatively stable and when reasonable attention coefficients are established, the major energy of the learning strategy turns into $\mathcal L_{cls}$ alternatively, as shown in Figure \ref{learning_strategy_curves} (b) and \ref{learning_strategy_curves} (c). This is defined as mid-term - from 100 to 1,700 steps.

At the tail of the training process, from epoch 1,700 to 2600, we observed that the learning curve was reasonably smooth and that $\gamma$ was stable and small. $1 - \gamma$ is a relatively large value, above 0.8 (one subtracts the value in Figure \ref{learning_strategy_curves} (a)), and the node classification loss thus draws greater attention from the optimizer. The optimizer adjusts the gradient almost based on the $\mathcal L_{cls}$ and therefore may neglect the subsequent impact on $\mathcal L_{att}$, which could result in a sharp increase in $\mathcal L_{att}$. As shown in Figure \ref{learning_strategy_curves} (c), we captured these fluctuations four times.
We therefore introduce the saturation parameter $\gamma_{str}$ to resist the sudden increase in $\gamma$. In this way, the fluctuations are handled appropriately, and better performance is achieved. Importantly, this type of fluctuation only occurs at the end of the training process, and it is not a necessary appearance for each training procedure.

In Figure \ref{attention_distribution}, another interesting point is that during the first 40 epochs, $\gamma$ is relatively large, and the HopGAT thus pays more attention to the adjustment of $\mathcal L_{att}$ and impacts the distribution of the attention coefficients. We can observe the emergence of the three separate clusters in this phase. After the 40 epochs, $\gamma$ starts to be restricted, and the boundaries of the clusters become more clear. This provides evidence that once reasonable supervision of the attention coefficients in the early phase is applied, the subsequent learning of the node classification can be jointly performed.

\section{Conclusion and Future Work}
This paper proposes a hop-aware attention supervision mechanism for the node classification task. Different from the previous works, we consider the influence of the hop values between a center and its neighbor nodes. Furthermore we jointly supervise the hop-aware attention coefficients and node classification error in the loss function, by which the loss function could be trained from more information of the context nodes. This method achieves state-of-the-art classification performance. In particular, it seems more effective for the inductive task in a graph with very few labeled nodes.

In the future, there are two interesting works can be done: (1) exploring a general hop-aware model which not only performs on node classification but also on link prediction task. (2) exploring the application of hop value and attention supervision mechanism on the heterogeneous graph network.







%
%
%
\newpage
\bibliographystyle{elsarticle-num}
\bibliography{paper}

\end{document}